\documentclass[twoside,leqno,twocolumn]{article}
\usepackage{ltexpprt}
\usepackage{multirow}
\usepackage{amssymb}
\usepackage[hidelinks]{hyperref}
\usepackage{amsmath}
\usepackage{graphicx}

\DeclareMathOperator*{\minimize}{minimize}

\begin{document}

\title{\Large Deep Transfer Reinforcement Learning for Text Summarization}
\author{Yaser Keneshloo \\ Discovery Analytics Center \\ Virginia Tech \\ yaserkl@vt.edu
\and
Naren Ramakrishnan \\ Discovery Analytics Center \\ Virginia Tech \\ naren@vt.edu
\and
Chandan K. Reddy \\ Discovery Analytics Center \\ Virginia Tech \\ reddy@cs.vt.edu
}

\date{}

\maketitle


\fancyfoot[R]{\scriptsize{Copyright \textcopyright\ 2019 by SIAM\\
Unauthorized reproduction of this article is prohibited}}





\begin{abstract}
\small\baselineskip=9pt Deep neural networks are data hungry models and thus face difficulties when attempting to train on small text datasets.
Transfer learning is a potential solution but their effectiveness in the text domain is not as explored as in areas
such as image analysis.
In this paper, we study the problem of transfer learning for text summarization and discuss why existing state-of-the-art models fail to generalize well on other (unseen) datasets.
We propose a reinforcement learning framework based on a self-critic policy gradient approach which achieves good generalization and state-of-the-art results on a variety of datasets.
Through an extensive set of experiments, we also show the ability of our proposed framework to fine-tune the text summarization model using only a few training samples.
To the best of our knowledge, this is the first work that studies transfer learning in text summarization and provides a generic solution that works well on unseen data.
\end{abstract}

\vspace{-3mm}
\section{Introduction}
\label{section:intro}
Text summarization is the process of summarizing a long document into few sentences that capture the essence of the whole document. 
In recent years, researchers have used news article datasets e.g., from CNN/DM~\cite{hermann2015teaching} and Newsroom~\cite{Grusky2018Newsroom} as a main resource for building and evaluating text summarization models. 
However, all these models suffer from a critical problem: \textit{a model trained on a specific dataset works well only on that dataset.} For instance, if a model is trained on the CNN/DM dataset and tested on the Newsroom dataset, the result is much poorer than when it is trained directly on the Newsroom dataset. 
This lack of generalization ability for current state-of-the-art models is the main motivation for our work.

This problem arises in situations where there is a need to perform summarization on a specific dataset, but either no ground-truth summaries exist for this dataset or where collecting ground-truth summaries could be expensive and time-consuming.
Thus, the only recourse in such a situation would be to simply apply a pre-trained summarization model to generate summaries for this data.
However, as discussed in this paper, this approach will fail to satisfy the basic requirements of this task and thus fails to generate high quality summaries.
Throughout our analysis, we work with two of the well-known news-related datasets for text summarization and one could expect that a model trained on either one of the datasets should perform well on the other or any news-related dataset.
On the contrary, as shown in Table~\ref{table:generalization_problem}, the Fast-RL model~\cite{chen2018fast} trained on CNN/DM, which holds the state-of-the-art result for text summarization task on the CNN/DM test dataset with 41.18\% a F-score according to the ROUGE-1 measure, will reach only a 21.93\% on this metric on the Newsroom test data, a performance fall of almost 20\%.
This observation shows that these models suffer from poor generalization capability.

\begin{figure}
    \centering
    \includegraphics[width=0.99\columnwidth]{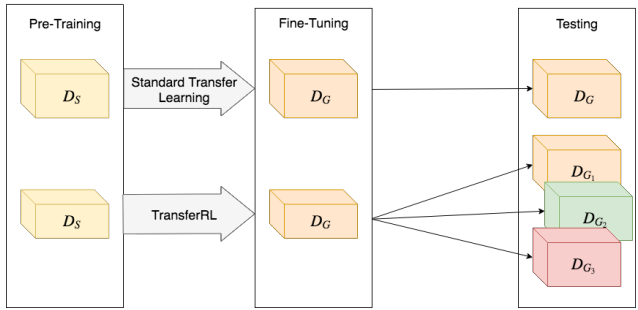}
    \caption{In standard transfer learning settings, a model is pre-trained on $D_S$, all network layers are transferred, the model is fine-tuned using $D_G$, and finally tested (only) on $D_G$. 
    On the contrary, our proposed method (TransferRL) uses $D_G$ to create a model that works well on a variety of (test) datasets.}
    \label{fig:tlvstrl}
\end{figure}

In this paper, we first study the extent to which the current state-of-the-art models are vulnerable in generalizing to other datasets and discuss how transfer learning could help in alleviating some of these problems.
In addition, we propose a solution based on reinforcement learning which achieves good generalization performance on a variety of summarization datasets.
Traditional transfer learning usually works by pre-training a model using a large dataset, fine-tuning it on a target dataset, and then testing the result on that target dataset.
However, our proposed method, as shown in Fig~\ref{fig:tlvstrl}, is able to achieve good results on a variety of datasets by only fine-tuning the model on a single dataset.
Therefore, it removes the requirement of training separate transfer models for each dataset.
To the best of our knowledge, this is the first work that studies transfer learning for the problem of text summarization and provides a solution for rectifying the generalization issue that arises in current state-of-the-art summarization models.
In addition, we conduct various experiments to demonstrate the ability of our proposed method to obtain state-of-the-art results on datasets with small amounts of ground-truth data.

The rest of the paper is organized as follows: Section~\ref{section:relatedwork} describes transfer learning methods and recent research related to this problem. Section~\ref{section:propsed} presents our proposed model for transfer learning in text summarization.
Section~\ref{section:exp} shows our experimental results and compares them with various benchmark datasets; Section~\ref{section:conclusion} concludes our discussion.

\begin{table}[!t]
\caption{The Pointer-Generator~\cite{see2017get} and Fast-RL~\cite{chen2018fast} models are trained using the CNN/DM dataset and tested on the CNN/DM and Newsroom datasets.}
\label{table:generalization_problem}
\centering
\scriptsize
\begin{tabular}{|c|c|c|c|c|c|c|}
\hline
                                                                   & \multicolumn{3}{c|}{\textbf{Pointer Generator}~\cite{see2017get}} & \multicolumn{3}{c|}{\textbf{Fast-RL~\cite{chen2018fast}}} \\ \hline
ROUGE                                                              & 1                           & 2                           & L                           & 1                         & 2                        & L              \\ \hline
\textbf{CNN/DM}                                                             & 39.77                       & 17.15                       & 36.18                       & 41.18            & 18.19           & 38.78        \\ \hline
\textbf{Newsroom}                                                           & 26.59                       & 14.09                       & 23.44                       & 21.93                     & 9.37                     & 19.61                    \\ \hline
\end{tabular}
\end{table}

\vspace{-3mm}
\section{Related Work}
\label{section:relatedwork}
Recently, there has been a surge in the development of deep learning based methods for building models that have the ability to transfer and generalize to other similar problems. 
Transfer learning (TL) has been well-studied in the domain of image processing; however its utility in NLP problems is yet to be thoroughly investigated.
In this section, we will review some of these works.

\vspace{-3mm}
\subsection{Transferring Trained Models.}
In these works, the underlying model is first trained on a specific dataset and then used as a pre-trained model for another problem or dataset. 
In this method, depending on the underlying model, one can transform different types of neural network layers from the pre-trained model to the transfer model. 
Examples of these transferable layers are the word embedding layer, the convolutional layers in a CNN model, Fully Connected (FC) hidden layers, and finally the output layer~\cite{ponti2018adversarial}.
Yosinski \textit{et al.}~\cite{yosinski2014transferable} studied the effect of transferring different layers of a deep neural network and found that lower-level layers learn general features while higher level layers capture mostly the specific characteristic of the problem at hand. 
Researchers have also demonstrated how one can transfer both low-level and high-level neural layers from a CNN for TL~\cite{donahue2014decaf}.

Recently, Semwal \textit{et al.}~\cite{semwal2018practitioners} used this idea of transferring various network layers for text classification. Aside from transferring the network layers, they also experimented with freezing or fine-tuning these layers after the transfer and concluded that fine-tuning the transfer layers will always provide a better result. Moreover, TL has also been studied in the context of the named entity recognition problem~\cite{sachan2017effective,lin2018neural}.
Our proposed method falls into this category. 
We not only study the effect of transferring network layers, but also propose a new co-training model for training text summarization models using reinforcement learning techniques.

\vspace{-3mm}
\subsection{Knowledge Distillation.}
Knowledge distillation refers to a class of techniques that trains a small network by transferring knowledge from a larger network.
These techniques are typically used when we require building models for devices with limited computational resources~\cite{ba2014deep}.
Usually, in these models, there is a teacher (larger model), a student (smaller model), and the goal is to transfer knowledge from teacher to student. 
Recently, researchers have also used this idea to create models using meta-learning~\cite{santoro2016meta}, few-shot learning~\cite{ravi2016optimization,snell2017prototypical}, one-shot learning~\cite{duan2017one,bertinetto2016learning}, and domain adaptation~\cite{ganin2016domain,tzeng2017adversarial}, mostly for image classification problems. 
However, the effect of these types of models on NLP tasks is yet to be studied or not well studied.

\vspace{-3mm}
\subsection{Building Generalized Models.}
Recently, McCann \textit{et al.}~\cite{mccann2018natural} released a challenge called Decathlon NLP which aims at solving ten different NLP problems with a single unified model.
The main intuition behind this model is to comprehend the impact of transferring knowledge from different NLP tasks on building a generalized model that works well on every task.
Although this model outperforms some of the state-of-the-art models in specific tasks, it fails to even reach baseline results in tasks like text summarization. 
We also observe such poor results from other generalized frameworks such as Google's Tensor2Tensor framework~\cite{vaswani2018tensor2tensor}.

\vspace{-3mm}
\subsection{Text Summarization.}
There is a vast amount of research work on the topic of text summarization using deep neural networks~\cite{shi2018neural}.
These works range from fully extractive methods~\cite{chen2018fast,narayan2018ranking,zhou2018neural} to completely abstractive ones~\cite{see2017get,kryscinski2018improving,gehrmann2018bottom}.
As one of the earliest works on using neural networks for extractive summarization, Nallapati \textit{et al.}~\cite{nallapati2017summarunner} proposed a framework that used a ranking technique to extract the most salient sentence in the input.
On the other hand, for abstractive summarization, it was Rush \textit{et al.}~\cite{rush2015neural} that for the first time used attention over a sequence-to-sequence (seq2seq) model for the problem of headline generation. 
To further improve the performance of these models, the pointer-generator model~\cite{nallapati2016abstractive} was proposed for successfully handling Out-of-Vocabulary (OOV) words.
This model was later improved by using the coverage mechanism~\cite{see2017get}.
However, all these models suffer from a common problem known as \textit{exposure bias} which refers to the fact that, during training, the model is trained by feeding ground-truth input at each decoder step, while during the test, it should rely on its own output to generate the next token.
Also, the training is typically done using cross-entropy loss, while during test, metrics such as ROUGE~\cite{lin2004rouge} or BLEU~\cite{papineni2002bleu} are used to evaluate the model. To tackle this problem, researchers suggested various models using scheduled sampling~\cite{bengio2015scheduled} and reinforcement learning based approaches~\cite{paulus2017deep,chen2018fast}.

Recently, several authors have investigated methods which try to first perform extractive summarization by selecting the most salient sentences within a document using a classifier and then apply a language model or a paraphrasing model~\cite{li2017paraphrase} on these selected sentences to obtain the final abstractive summarization~\cite{chen2018fast,narayan2018ranking,zhou2018neural}. However, none of these models, as discussed in this paper (and shown in Table~\ref{table:generalization_problem}) have the capability to generalize to other datasets and thus only perform well for the specific dataset used as target data during the pre-training process.
\vspace{-3mm}
\section{Proposed Model}
\label{section:propsed}
In this paper, we propose various transfer learning methods for the problem of text summarization. 
For all experiments, we consider two different datasets: $D_S$, a \textit{source dataset} used to train the pre-trained model, while $D_G$, the
\textit{target dataset}\footnote{Note that, we use $D_G$ instead of $D_T$ for the target dataset to avoid confusing this $T$ subscript with time.}, is the dataset used to fine-tune our pre-trained model.
Following the idea of transferring layers of a pre-trained model, our first proposed model transfers different layers of a pre-trained model trained using $D_S$ and fine-tunes them using $D_G$. 
We then propose another method which uses a novel reinforcement learning (RL) framework to train the transfer model using training signals received from both $D_S$ and $D_G$.

\vspace{-3mm}
\subsection{Transferring Network Layers.}
There are various network layers used in a deep neural network.
For instance, if the model has a CNN encoder and a LSTM decoder, the CNN layers and the hidden decoder layers trained on $D_S$ could be used to fine-tune using $D_G$. 
Moreover, the word embedding representation is a key layer in such a model.
Either, we use pre-trained word embeddings such as Glove~\cite{pennington2014glove} during the training of $D_S$ or let $D_S$ drive the inference of its own word embeddings. 
We can still let the model to fine-tune a pre-trained word embedding during the training of such a model.
In summary, we can transfer the embedding layer, convolutional layer (if using CNN), hidden layers (if using LSTM), and the output layer in a text summarization transfer learning problem.
One way to understand the effect of each of these layers is to fine-tune or freeze these layers during model transfer and report the best performing model.
However, as suggested by~\cite{semwal2018practitioners}, the best performance is realized when all layers of a pre-trained model on $D_S$ are transferred and the model is led to fine-tune itself using $D_G$. 
Therefore, we follow the same practice and let the transferred model fine-tune all trainable variables in our model.
As shown later in the experimental result section, this way of transferring network layers provides a strong baseline in text summarization and the performance of our proposed reinforced model is close to this baseline.
However, one of the main problems with this approach is that, not only should source dataset $D_S$ should contain a large number of training samples but $D_G$ must also have a lot of training samples to be able to fine-tune the pre-trained model and generalize the distribution of the pre-trained model parameters.
Therefore, a successful transfer learning using this method requires a large number of samples both for $D_S$ and $D_G$.
This could be problematic, specifically for cases where the target dataset is small and fine-tuning a model will cause over-fitting.
For these reasons, we will propose a model which uses reinforcement learning to fine-tune the model only based on the reward that is obtained over the target dataset.

\vspace{-3mm}
\subsection{Transfer Reinforcement Learning (TransferRL).}
In this section, we explain our proposed reinforcement learning based framework for knowledge transfer in text summarization. The basic underlying summarization mechanism used in our work is the pointer-generator model~\cite{see2017get}.

\vspace{-3mm}
\subsubsection{Why Pointer-Generator?}
The reason we choose a pointer-generator model as the basis of our framework is its ability to handle Out-of-Vocabulary (OOV) words which is necessary for transfer learning.
Note that once a specific vocabulary generated from $D_S$ is used to train the pre-trained model, we cannot use a different vocabulary set during the fine-tuning stage on $D_G$, since the indexing of words could change for words in the second dataset\footnote{For instance, the word ``is'' could have index 1 in the first dataset while it could have index 10 in the second dataset.}.
According to our experiments, amongst the top 50K words in the CNN/DM and Newsroom datasets, only 39K words are common between the two datasets and thus a model trained on each of these datasets will have more than 11K OOVs during the fine-tuning step.
Therefore, a framework that is not able to handle these OOV words elegantly will demonstrate significantly poor results after the transfer.
One na\"ive approach in resolving this problem could be to use a shared set of vocabulary words between $D_S$ and $D_G$.
However, such a model will still suffer from inability to generalize to other datasets with a different vocabulary set.

\vspace{-3mm}
\subsubsection{Pointer-Generator.}
\begin{figure}[!t]
    \centering
    \includegraphics[width=0.99\columnwidth]{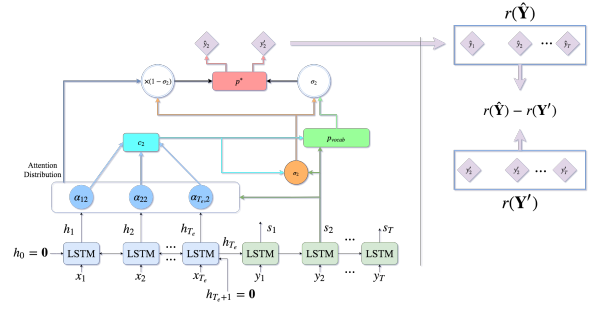}
    \caption{Pointer-generator w. self-critic policy gradient}
    \label{fig:pointer_generator}
\end{figure}

As shown in Fig~\ref{fig:pointer_generator}, a pointer-generator model comprises of a series of LSTM encoders (blue boxes) and LSTM decoders (green boxes). Let us consider dataset $D=\{d_1, \cdots, d_N\}$ as a dataset that contains $N$ documents along with their summaries. Each document is represented by a series of $T_e$ words, i.e. $d_i = \{x_1, \cdots, x_{T_e}\}$, where $x_t\in V=\{1,\cdots, |V|\}$. Each encoder takes the embedding of word $x_t$ as the input and generates the output state $h_t$.
The decoder, on the other hand, takes the last state from the encoder, i.e., $h_{T_e}$, and starts generating an output of size $T<T_e$, $\hat{\textbf{Y}}=\{\hat{y}_1,\hat{y}_2,\cdots,\hat{y}_T\}$, based on the current state of the decoder $s_t$, and the ground-truth summary word $y_t$. At each step of decoding $j$, the attention vector $\alpha_j$, context vector $c_j$, and output distribution $p_{vocab}$ can be calculated as follows:
\begin{equation}
\begin{array}{lcl}
    f_{ij} &=& v_1^T tanh(W_h h_i + W_s s_j + b_1) \\
    \alpha_j &=& softmax(f_{j}) \\ 
    c_j &=& \sum_{i=1}^{T_e} \alpha_{ij} h_i \\
    p_{vocab} &=& softmax(v_2 (v_3 [s_j\oplus c_j] + b_2) + b3)
\end{array}
\end{equation}
where $v_{1,2,3}$, $b_{1,2,3}$, $W_h$, and $W_s$ are trainable model parameters and $\oplus$ is the concatenation operator. In a simple sequence-to-sequence model with attention, we use $p_{vocab}$ to calculate the cross-entropy loss. However, since $p_{vocab}$ only captures the distribution of words within the vocabulary, this will generate a lot of OOV words during the decoding step. A pointer-generator model mitigates this issue by using a switching mechanism which either chooses a word from the vocabulary with a certain probability $\sigma$ or from the original document using the attention distribution with a probability of (1-$\sigma$) as follows:
\begin{equation}
\begin{array}{lcl}
    \sigma_j &=& (W_c c_j + W_s s_j + W_x x_j + b_4) \\
    p^*_j &=& \sigma_j p_{vocab} + (1-\sigma_j)\sum_{i=1}^{T_e} \alpha_{ij} 
\end{array}
\label{eq:pointer}
\end{equation}
where $W_c$, $W_x$, and $b_4$ are trainable model parameters and if a word $x_j$ is OOV, then $p_{vocab}=0$ and the model will rely on the attention values to select the right token. Once the final probability is calculated using Eq.~(\ref{eq:pointer}), the cross-entropy (CE) loss is calculated as follows:
\begin{equation}
\mathcal{L}_{CE}=-\sum_{t=1}^{T}\log{p^*_{\theta}(y_{t}|e(y_{t-1}),s_t, c_{t-1},\textbf{X})}
\label{eq:cel}
\end{equation}
where $\theta$ shows the training parameters and $e(.)$ returns the word embedding of a specific token. However, as mentioned in Section~\ref{section:relatedwork}, one of the main problems with cross-entropy loss is the exposure bias ~\cite{bengio2015scheduled,paulus2017deep} which occurs due to the inconsistency between the decoder input during training and test. A model that is trained using only CE loss does not have the generalization power required for transfer learning, since such a model is not trained to generate samples from its own distribution and heavily relies on the ground-truth input. Thus, if the distribution of input data changes (which can likely happen during transfer learning on another dataset), the trained model will have to essentially re-calibrate every transferred layer to achieve a good result on the target dataset. To avoid these problems, we propose a reinforcement learning framework which slowly removes the dependency of model training on the CE loss and increases the reliance of the model on its own output.

\vspace{-3mm}
\subsubsection{Reinforcement Learning Objective.}
In RL training, the focus is on minimizing the negative expected reward rather than on directly minimizing the CE loss.
This allows the framework to not only use the model's output for training itself, but also helps in training the model based on the metric that is used during decoding (such as ROUGE).
To achieve this, during RL training, the following objective is minimized:
\begin{equation}
\minimize\ \mathcal{L}_{RL} =-\mathop{\mathbb{E}_{y^{\prime}_1,\cdots,y^{\prime}_T \sim {p^{*}_{\theta}(y^{\prime}_1,\cdots,y^{\prime}_T)}}}[r(y^{\prime}_1,\cdots,y^{\prime}_T)]
\label{eq:pg}
\end{equation}
where ${y^{\prime}_1, \cdots, y^{\prime}_T}$ are sample tokens drawn from the output of the policy ($p_{\theta}$), i.e., $p^{*}_1, \cdots, p^{*}_T$.
In practice, we usually sample only one sequence of tokens to calculate this expectation.
Hence, the derivative of the above loss function is given as follows:
\begin{equation}
\footnotesize
\nabla_\theta \mathcal{L}_{RL} = -\mathop{\mathbb{E}}_{y^{\prime}_1,\cdots,y^{\prime}_T\sim p^{*}_\theta}[\nabla_\theta\log{p^{*}_\theta(y^{\prime}_1,\cdots,y^{\prime}_T)}r(y^{\prime}_1,\cdots,y^{\prime}_T)]
\end{equation}

This minimization can be further improved by adding a baseline reward.
In text summarization, the baseline reward could either come from a separate network called a critic network~\cite{chen2018fast} or it could be the reward from a sequence coming from greedy selection over $p^{*}_t$~\cite{paulus2017deep}.
In this work, we consider the greedy sequence as the baseline.
In summary, the objective that we minimize during RL training is as follows:
\begin{equation}
\begin{array}{cl}
\mathcal{L}_{RL} =& \sum_t -\log{p^{*}_\theta(y_{t}|y^{\prime}_{t-1}, s_{t},c_{t-1},\textbf{X})}\times\\
&\Big(r(\hat{y}_{1},\cdots,\hat{y}_{T}) - r(y^{\prime}_{1},\cdots,y^{\prime}_{T})\Big)
\end{array}
\label{eq:scloss}
\end{equation}
where $\hat{y}_t$ represents the greedy selection at time $t$.
This model is also known as a \textit{self-critic policy gradient} approach since the model uses its own greedy output to create the baseline.
Moreover, the model uses the sampled sentence as the target for training rather than the ground-truth sentence. Therefore, given this objective, the model focuses on samples that do better than greedy selection during training while penalizing those which do worse than greedy selection.

\begin{figure}
    \centering
    \includegraphics[scale=0.25]{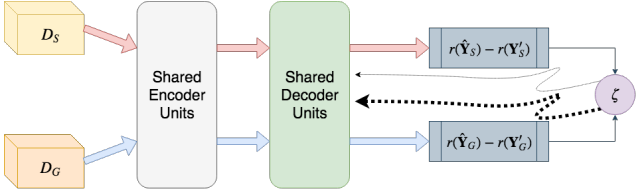}
    \caption{The proposed TransferRL framework.
    The encoder and decoder units are shared between the source ($D_S$) and target datasets ($D_G$).}
    \label{fig:transferrl}
\end{figure}

\vspace{-3mm}
\subsubsection{Transfer Reinforcement Learning.}
Although a model trained using Eq.~(\ref{eq:scloss}) does not suffer from exposure bias, it can still perform poorly in a transfer learning setting.
This is mostly due to the fact that
the model
is still being trained using the distribution from the source dataset and once transferred to the target dataset, aims to generate samples according to the distribution of the source dataset. Therefore, we need a model that not only remembers the distribution of the source dataset but also tries to learn and adapt to the distribution of the target dataset. The overall RL-based framework that is being proposed in this paper is shown in Fig.~\ref{fig:transferrl}.
At each step, we select a random mini-batch from $D_S$ and $D_G$ and feed them to the shared encoder units and the decoder starts decoding for each mini-batch. 
Once the decoding is completed, the model generates a sentence based on greedy selection and another by sampling from the output distribution.
Finally, we calculate the TransferRL loss according to Eq.~(\ref{eq:trl}) and back-propagate the error according to the trade-off parameter $\zeta$.
The thick and thin dashed lines in this plot shows the effect of $\zeta$ on the extent to which the model needs to rely on either $D_S$ or $D_G$ for back-propagating the error.

Let us consider sequences drawn from greedy selection and sampling from the source dataset $D_S$ and the target dataset $D_G$ as $\hat{\textbf{Y}}_S$, $\textbf{Y}^{\prime}_S$, $\hat{\textbf{Y}}_G$, and $\textbf{Y}^{\prime}_G$, respectively. We will define the transfer loss function using these variables as follows:
\begin{equation}
\footnotesize
\begin{array}{clc}
    \mathcal{L}_{TRL} = -\sum_{t=1}^{T}
    \Bigg(&(1-\zeta) \log{p^{*}_{\theta}(y^{S}_{t}|\textbf{U}_S)} \Big( r(\hat{\textbf{Y}}_S)-r(\textbf{Y}^{\prime}_S)\Big)+&\\
    &\zeta \log{p^{*}_{\theta}(y^{G}_{t}|\textbf{U}_G)} \Big(r(\hat{\textbf{Y}}_G)-r(\textbf{Y}^{\prime}_G)\Big)\Bigg)&
\end{array}
\label{eq:trl}
\end{equation}
where $\textbf{U}_{S}=\{e_{S}(y^{\prime}_{S,t-1}),s_t, c_{t-1},\textbf{X}_{S}\}$, $\textbf{U}_{G}=\{e_{S}(y^{\prime}_{G,t-1}),s_t, c_{t-1},\textbf{X}_{G}\}$, and $\zeta\in [0,1]$ controls the trade-off between self-critic loss of the samples drawn from the source dataset and from the target dataset.
Therefore, a $\zeta=0$ means that we train the model only using the samples from the source dataset, while $\zeta=1$ means that model is trained only using samples from the target dataset. 
As seen in Eq.~(\ref{eq:trl}), the decoder state $s_t$ and the context vector $c_{t-1}$ are shared between the source and target samples.
Moreover, we use a shared embedding trained on the source dataset, $e_{S}(.)$ for both datasets while the input data given to the encoder, i.e., $X_{S}$ and $X_{G}$, come from source and target datasets.

In practice, the RL objective loss only activates after a good pre-trained model is obtained.
We follow the same practice and first pre-train the model using the source dataset and then activate the transfer RL loss in Eq.~(\ref{eq:trl}) by combining this loss with the CE loss from Eq.~(\ref{eq:cel}) using the parameter $\eta\in[0,1]$ as follows:
\begin{equation}
    \mathcal{L}_{Mixed}= (1-\eta) \mathcal{L}_{CE} + \eta \mathcal{L}_{TRL}
\end{equation}

\section{Experimental Results}
\label{section:exp}
We performed a range of experiments to understand the dynamics of transfer learning and to investigate best practices for obtaining a generalized model for text summarization. In this section, we discuss some of the insights we gained through our experiments. All evaluations are done using ROUGE 1, 2, and L F-scores on the test data. All our ROUGE scores have a 95\% confidence interval of $\pm 0.25$ as reported by the official ROUGE script\footnote{\url{https://pypi.org/project/pyrouge/}}. Similar to multi-task learning frameworks such as DecaNLP~\cite{mccann2018natural}, we use a measure for comparing the result of transfer learning on various datasets by taking the average score of each measure over these datasets. In addition, we also introduce a weighted average score which takes into account the size of each dataset as the weight for averaging the values\footnote{Note that, due to page limitations, some of the results are presented in the Arxiv version of the paper \url{https://arxiv.org/abs/1810.06667}.}.

\vspace{-2mm}
\subsection{Datasets.}
We use four widely used datasets in text summarization for our experiments.
The first two datasets are Newsroom~\cite{Grusky2018Newsroom} and CNN/Daily Mail~\cite{hermann2015teaching} which are used for training our models, while the DUC 2003 and DUC 2004 datasets are only used to test the generalization capability of each model.
Table~\ref{table:dsstat} shows some of the basic statistics of these datasets.
In all these datasets, a news article is accompanied by 1 to 4 human-written summaries and, therefore, will cover a wide range of challenges for transfer learning. For instance, a model that is trained on the Newsroom dataset will most likely generate only one long summary sentence, while for the CNN/DM dataset, the model is required to generate up to four smaller summary sentences. For all experiments, we either use Newsroom as $D_S$ and CNN/DM as $D_G$, or vice-versa.

\begin{table}[!t]
\centering
\scriptsize
\caption{Basic statistics for the datasets used in our experiments.}
\label{table:dsstat}
\begin{tabular}{|l|c|c|c|c|}
\hline
                                                                                       & \textbf{Newsroom} & \textbf{CNN/DM} & \textbf{DUC'03} & \textbf{DUC'04} \\ \hline
\textbf{\# Train}                                                                      & 994,001            & 287,226          & 0               & 0               \\ \hline
\textbf{\# Val}                                                                       & 108,312            & 13,368           & 0               & 0               \\ \hline
\textbf{\# Test}                                                                       & 108,655            & 11,490           & 624             & 500             \\ \hline
\textbf{\begin{tabular}[c]{@{}c@{}}Avg. \#\\ summary\\ sentences\end{tabular}}      & 1.42              & 3.78            & 4               & 4               \\ \hline
\textbf{\begin{tabular}[c]{@{}c@{}}Avg. \# \\words in\\ summary\end{tabular}} & 21                & 14.6            & 11.03           & 11.43           \\ \hline
\end{tabular}
\end{table}

\vspace{-2mm}
\subsection{Training Setup.}
For each experiment, we run our model for 15 epochs during pre-training and 10 epochs during the transfer process, and an extra 2 epochs for the coverage mechanism.
We use a batch size of 48 during training, the encoder reads the first 400 words, and the decoder generates a summary with 100 words.
Both encoder and decoder units have a hidden size of 256 while the embedding dimension is set to 128 and we learn the word embedding during training.
For all models, we used the top 50K words in each dataset as the vocabulary and during test we use beam search of size 4.
We use AdaGrad to optimize all models with an initial learning rate of $\gamma_0=0.15$ during pre-training and $\gamma_0=0.001$ during RL and coverage and linearly decrease this learning rate based on the epoch numbers as $\gamma_t=\gamma_0/epoch$.
Moreover, $\zeta$ is set to zero at the start of RL training and is increased linearly so that it gets to 1 by the end of training.
During RL training, we use scheduled sampling with sampling probability equal to the $\zeta$ value. We use the RLSeq2Seq~\cite{keneshloo2018deep} framework to build our model.

\begin{table*}[!t]
\footnotesize
\centering
\caption{Results on Newsroom, CNN/DM, DUC'03, and DUC'04 test data. $D_S$ shows the dataset that is used during pre-training and $D_G$ is our target dataset. \textbf{N} stands for the Newsroom dataset and \textbf{C} denotes the CNN/DM dataset. The method column shows whether we use CE loss, transferring layers (TL), or TransferRL (TRL) loss during training.  We use a coverage mechanism for all experiments. The result from the proposed method is shown with $\star$.}
\label{table:trlresults}
\begin{tabular}{|c|c|c|c|c|c|c|c|c|c|c|c|c|c|c|c|c|}
\hline
\textbf{\#} & \textbf{$D_S$}    & \textbf{$D_G$}    & \multicolumn{2}{c|}{\textbf{Method}} & \multicolumn{3}{c|}{\textbf{Newsroom}}           & \multicolumn{3}{c|}{\textbf{CNN/DM}}             & \multicolumn{3}{c|}{\textbf{DUC'03}}             & \multicolumn{3}{c|}{\textbf{DUC'04}}             \\ \hline
\multicolumn{5}{|c|}{\textbf{}}                                & \textbf{$R_1$} & \textbf{$R_2$} & \textbf{$R_L$} & \textbf{$R_1$} & \textbf{$R_2$} & \textbf{$R_L$} & \textbf{$R_1$} & \textbf{$R_2$} & \textbf{$R_L$} & \textbf{$R_1$} & \textbf{$R_2$} & \textbf{$R_L$} \\ \hline
\textbf{1} & \multicolumn{2}{|c|}{\textbf{N}}        & \multicolumn{2}{c|}{CE Loss}            & 36.16          & 24.33          & 32.87          & 33.58          & 12.76          & 29.72          & 28.03          & 9.15           & 24.75             & 29.85             & 10.3          & 26.7           \\ \hline
\multicolumn{1}{|c|}{\textbf{2}} & \multicolumn{2}{|c|}{\textbf{C+N}} & \multicolumn{2}{c|}{CE Loss} & \multicolumn{1}{c|}{30.26} & \multicolumn{1}{c|}{17.68} & \multicolumn{1}{c|}{27.03} & \multicolumn{1}{c|}{\textbf{38.23}} & \multicolumn{1}{c|}{\textbf{16.31}} & \multicolumn{1}{c|}{\textbf{34.66}} & \multicolumn{1}{c|}{26.71} & \multicolumn{1}{c|}{7.81} & \multicolumn{1}{c|}{24.13} & \multicolumn{1}{c|}{27.96} & \multicolumn{1}{c|}{8.25} & \multicolumn{1}{c|}{25.25} \\ \hline
\textbf{3} & \textbf{N} & \textbf{C}          & \multicolumn{2}{c|}{TL}            & 35.37          & 23.45          & 32.07          & 34.51          & 13.49          & 30.61          & 28.19          & 9.34           & 24.96          & 29.83          & 9.98           & 26.66          \\ \hline
\textbf{4} & \textbf{N} & \textbf{C}  & \multicolumn{2}{c|}{TRL$^\star$}           & \textbf{36.5}  & \textbf{24.77} & \textbf{33.25} & 35.24          & 13.56          & 31.33          & \textbf{28.46} & \textbf{9.65}  & \textbf{25.45} & \textbf{30.45} & \textbf{10.63} & \textbf{27.42} \\ \hline
\end{tabular}
\end{table*}

\begin{table*}[!t]
\centering
\footnotesize
\caption{Normalized and weighted normalized ROUGE F-Scores for Table~\ref{table:trlresults}.}
\label{table:ntrlresults}
\begin{tabular}{|c|c|c|c|c|c|c|c|c|c|}
\hline
\textbf{\#} &\textbf{$D_S$}    & \textbf{$D_G$}    & \multicolumn{1}{c|}{\textbf{Method}} & \multicolumn{3}{c|}{\textbf{Avg. Score}}  & \multicolumn{3}{c|}{\textbf{\begin{tabular}[c]{@{}c@{}}Weighted\\ Avg. Score\end{tabular}}}\\ \hline
\multicolumn{4}{|c|}{\textbf{}}                                & \textbf{$R_1$} & \textbf{$R_2$} & \textbf{$R_L$} & \textbf{$R_1$} & \textbf{$R_2$} & \textbf{$R_L$} \\ \hline
\textbf{1}  & \multicolumn{2}{|c|}{\textbf{N}}        & \multicolumn{1}{c|}{CE Loss}            & 31.91          & 14.14          & 28.51          & 35.58                         & 21.73                        & 32.16                        \\ \hline
\multicolumn{1}{|c|}{\textbf{2}}  & \multicolumn{2}{|c|}{\textbf{C+N}} & \multicolumn{1}{c|}{CE Loss} & \multicolumn{1}{c|}{30.79} & \multicolumn{1}{c|}{12.51} & \multicolumn{1}{c|}{27.77} & \multicolumn{1}{c|}{32.04} & \multicolumn{1}{c|}{17.36} & \multicolumn{1}{c|}{28.74} \\ \hline
\textbf{3}  & \textbf{N} & \textbf{C}         & \multicolumn{1}{c|}{TL}            & 31.98          & 14.07          & 28.58          & 35.17                         & 21.21                        & 31.74                        \\ \hline
\textbf{4}  & \textbf{N} & \textbf{C} & \multicolumn{1}{c|}{TRL$^\star$}           & \textbf{32.66} & \textbf{14.65} & \textbf{29.36} & \textbf{36.21}                & \textbf{22.25}               & \textbf{32.81}               \\ \hline
\end{tabular}
\end{table*}

\vspace{-2mm}
\subsection{Effect of Dataset Size.}
We will now discuss some of the insights we gained starting with understanding the effect of data size for pre-training.
According to our experiments (as shown in Table~\ref{table:ntrlresults}), on average, a model trained using the Newsroom dataset as the source dataset $D_S$ has much better performance than models that use CNN/DM as the $D_S$ in almost all configurations\footnote{Due to space constraints, we only report the results from this setup and refer the readers to the Arxiv version of the paper.
}.
This is not a surprising result since deep neural networks are data hungry models and typically work the best when provided with a large number of samples. The first experiment in Table~\ref{table:trlresults} and Table~\ref{table:ntrlresults} uses only the Newsroom dataset for training the model and not surprisingly it performs good on this dataset; however as discussed earlier, its performance on other datasets is poor.

\vspace{-2mm}
\subsection{Common Vocabulary.}
As mentioned in Section~\ref{section:propsed}, one way to avoid excessive OOV words in transfer learning between two datasets is to use a common vocabulary between $D_S$ and $D_G$ and train a model using this common vocabulary set.
Although a model trained using such a vocabulary could perform well on these two datasets, it still suffers from poor generalization to other unseen datasets.
To demonstrate this, we combine all articles in CNN/DM and Newsroom training datasets to create a single unified dataset (C+N in Table~\ref{table:trlresults} and Table~\ref{table:ntrlresults}) and train a model using CE loss in Eq.~(\ref{eq:cel}) and the common set of vocabulary. The result of this experiment is shown as Experiment 2 in Table~\ref{table:trlresults} and in Table~\ref{table:ntrlresults}.
As shown here, by comparing these results to Experiment 1, we see that combining these two datasets will decrease the performance on Newsroom, DUC'03, and DUC'04 test datasets but will increase the performance for CNN/DM test data.
Moreover, by comparing the generalization ability of this method on DUC'03 and DUC'04 datasets, we see that it performs up to 2\% worse than the proposed method.
This is also witnessed by comparing the average scores and weighted average scores of our proposed model against this model.
On average, our method improves up to 4\% compared to this method according to the $R_1$ weighted average score.

\vspace{-2mm}
\subsection{Transferring Layers.}
\label{section:exp:layertransfer}
In this experiment, we discuss the effect of transferring different layers of a pre-trained model for transfer learning.
In the pointer-generator model described in Section~\ref{section:propsed}, the embedding matrix, the encoder and decoder model parameters are the choices for layers we can use for transfer learning.
For this experiment, we pre-train our model using $D_S$ and during transfer learning, we replace $D_S$ with $D_G$ and continue training of the model with CE loss.
As shown in Tables~\ref{table:trlresults} and~\ref{table:ntrlresults} this way of transferring network layers provides a strong baseline for comparing the performance of our proposed method.
These results show that even a simple transferring of layers could provide enough signals for the model to adapt itself to the new data distribution.
However, as discussed earlier in Section~\ref{section:propsed}, this way of transfer learning tends to completely forget the pre-trained model distribution and entirely changes the final model distribution toward the dataset used for fine-tuning.
This effect can be observed in Table~\ref{table:trlresults} by comparing the result of experiments 1 and 3.
As shown in this table, after transfer learning the performance drops on the Newsroom test dataset (from 36.16 to 35.37 based on $R_1$) while it increases on the CNN/DM dataset (from 33.58 to 34.51 based on $R_1$).
However, since our proposed method tries to remember the distribution of the pre-trained model (through the $\zeta$ parameter) and slowly changes the distribution of the model according to the distribution coming from the target dataset, it performs better than simple transfer learning on these two datasets.
This is shown by comparing the result in experiments 3 and 4 in Table~\ref{table:trlresults}, which shows that our proposed model performs better than na\"ive transfer learning in all test datasets.



\vspace{-2mm}
\subsection{Effect of Zeta.}
As mentioned in Section~\ref{section:propsed}, the trade-off between emphasizing the training to samples drawn from $D_S$ or $D_G$ is controlled by the hyper-parameter $\zeta$.
To see the effect of $\zeta$ on transfer learning, we clip the $\zeta$ value at 0.5 and train a separate model using this objective.
Basically, a $\zeta=0.5$ means that we treat the samples coming from source and target datasets equally during training.
Therefore, for these experiments, we start $\zeta$ at zero and increase it linearly till the end of training but clip the final $\zeta$ value at 0.5.
Table~\ref{table:zetaclipping} shows the result of this experiment. For simplicity sake, we provide the result of our proposed model achieved from not clipping $\zeta$ along with these results.
By comparing the results from these two setups, we can see that, on average, increasing the value of $\zeta$ to 1.0 will yield better results than clipping this value at 0.5.
For instance, according to the average and weighted average score there is an increase of 0.7\% in ROUGE-1 and ROUGE-L scores when we do not clip the $\zeta$ at 0.5.
By comparing the CNN/DM $R_1$ score in this table, we see that clipping the $\zeta$ value will definitely hurt the performance on $D_G$ since the model shows equal attention to the distribution coming from both datasets.
On the other hand, the surprising component here is that, by avoiding $\zeta$ clipping, the performance on the source dataset also increases~\footnote{For $\zeta \in (0.5,1)$, we have only seen small improvement in the results and hence we omitted them from this section.}.

\begin{table}[]
\centering
\caption{Result of TransferRL after clipping $\zeta$ at 0.5 and $\zeta=1.0$ on Newsroom, CNN/DM, DUC'03, and DUC'04 test datasets, along with the average and weighted average scores.}
\label{table:zetaclipping}
\scriptsize
\begin{tabular}{|c|c|c|c|c|c|c|}
\hline
$\zeta$                                                       & \multicolumn{3}{c|}{\textbf{0.5}} & \multicolumn{3}{c|}{\textbf{1.0}}                         \\ \hline
                                                              & $R_1$  & $R_2$  & $R_L$  & $R_1$          & $R_2$          & $R_L$          \\ \hline
\textbf{Newsroom}                                                      & 36.06  & 24.23  & 32.78  & \textbf{36.5}  & \textbf{24.77} & \textbf{33.25} \\ \hline
\textbf{CNN/DM}                                                        & 33.7   & 12.83  & 29.81  & \textbf{35.24} & \textbf{13.56} & \textbf{31.33} \\ \hline
\textbf{DUC'03}                                                        & 28.3   & 9.54   & 25.04  & \textbf{28.46} & \textbf{9.65}  & \textbf{25.45} \\ \hline
\textbf{DUC'04}                                                        & 29.88  & 10.23  & 26.8   & \textbf{30.45} & \textbf{10.63} & \textbf{27.42} \\ \hline
\begin{tabular}[c]{@{}c@{}}\textbf{Avg. Score}\end{tabular}          & 31.98  & 14.2   & 28.6   & \textbf{32.66} & \textbf{14.65} & \textbf{29.36} \\ \hline
\begin{tabular}[c]{@{}c@{}}\textbf{Weighted}\\\textbf{Avg. Score}\end{tabular} & 35.52  & 21.66  & 32.11  & \textbf{36.21} & \textbf{22.25} & \textbf{32.81} \\ \hline
\end{tabular}
\end{table}

\vspace{-2mm}
\subsection{Transfer Learning on Small Datasets.}
As discussed in Section~\ref{section:intro}, transfer learning is good for situations where the goal is to do summarization on a dataset with little or no ground-truth summaries.
For this purpose, we conducted another set of experiments to test our proposed model on transfer learning using DUC'03 and DUC'04 datasets as our target datasets, i.e., $D_G$.
For these experiments, we randomly pick 20\% of each dataset as our training set, 10\% as a validation dataset, and the rest of the dataset as our test data.
This will generate 124 and 100 articles as our training dataset for DUC'03 and DUC'04, respectively.
Similar to other experiments in this paper, we use CNN/DM and Newsroom as $D_S$\footnote{The CNN/DM results are excluded due to space constraints; the reader can refer to the Arxiv version of our paper.} and use DUC'03 and DUC'04 as $D_G$ during transfer learning.
Due to the size of these datasets, the models are trained only for 3000 iterations during fine-tuning and the best model is selected according to the validation set.
Tables~\ref{table:duc03} and~\ref{table:duc04} depict the results of this experiment.
As shown in these tables, for DUC'03, when we simply transfer network layers, it performs slightly better (not statistically higher according to a 95\% confidence interval) than our proposed model; however, our proposed model will achieve a far better result on DUC'04.
As shown in these tables, the results achieved from fine-tuning a pre-trained model using these datasets is very close to the ones achieved in Table~\ref{table:trlresults} and in the case of the DUC'04 dataset, our proposed method in Table~\ref{table:trlresults} achieves even better results than the ones shown in Table~\ref{table:duc04}.
This shows the ability of our proposed framework in generalizing to unseen datasets.
Note that, unlike these experiments, the proposed model in Table~\ref{table:trlresults} has no information about the data distribution of DUC'03 and DUC'04 and still performs better on these datasets.

\begin{table}[!t]
\footnotesize
\centering
\caption{Result of transfer learning methods using the Newsroom dataset for pre-training and DUC'03 for fine-tuning. The underlined result shows that the improvement from TL is not statistically significant compared to the proposed model.}
\label{table:duc03}
\begin{tabular}{|c|c|c|c|c|c|}
\hline
\textbf{$D_S$} & \textbf{$D_G$}    & \textbf{Method} & \textbf{$R_1$} & \textbf{$R_2$} & \textbf{$R_L$} \\ \hline
\textbf{N} & \textbf{DUC'03}        & TL            & \underline{28.9}          & \underline{9.73}           & \underline{25.54}          \\ \hline
\textbf{N} & \textbf{DUC'03} & TRL$^\star$           & 28.76          & 9.5           & 25.39          \\ \hline
\end{tabular}
\end{table}

\begin{table}[!t]
\footnotesize
\centering
\caption{Result of transfer learning using Newsroom for pre-training and DUC'04 for fine-tuning.}
\label{table:duc04}
\begin{tabular}{|c|c|c|c|c|c|}
\hline
\textbf{$D_S$} & \textbf{$D_G$}    & \textbf{Method} & \textbf{$R_1$} & \textbf{$R_2$} & \textbf{$R_L$} \\ \hline
\textbf{N} & \textbf{DUC'04}         & TL            & 27.68          & 9.09           & 25.29          \\ \hline
\textbf{N} & \textbf{DUC'04} & TRL$^\star$           & \textbf{29.54}          & \textbf{9.99}           & \textbf{26.56}          \\ \hline
\end{tabular}
\end{table}
\vspace{-2mm}
\subsection{Other Generalized Models.}
We also compare the performance of our proposed model against some of the recent methods from multi-task learning.
In text summarization, the DecaNLP~\cite{mccann2018natural} and Tensor2Tensor~\cite{vaswani2018tensor2tensor} are two of the most recent frameworks that use multi-task learning.
Following the setup in these works, we focus on models that are trained using CNN/DM datasets and report the average ROUGE 1, 2, and L F-scores for our best performing model.
Table~\ref{table:multitask} compares the result of our proposed approach against these methods.

\vspace{-2mm}
\begin{table}[!t]
\centering
\scriptsize
\caption{Comparing our best performing model with state-of-the-art multi-task learning frameworks on the CNN/DM dataset and according to the average of ROUGE 1, 2, and L F-scores. The result with $*$ is the same as reported in the original paper.}
\label{table:multitask}
\begin{tabular}{|c|c|c|c|}
\hline
\textbf{}              & \textbf{DecaNLP~\cite{mccann2018natural}} & \textbf{Tensor2Tensor~\cite{vaswani2018tensor2tensor}} & \textbf{\begin{tabular}[c]{@{}c@{}}Proposed\\ Model\end{tabular}} \\ \hline
\textbf{\begin{tabular}[c]{@{}c@{}}Average\\ ROUGE\end{tabular}} & $25.7^{*}$                                           & 27.4                                                 & \textbf{31.12}          \\ \hline
\end{tabular}
\end{table}
\vspace{-2mm}
\section{Conclusion}
\label{section:conclusion}
In this paper, we tackled the problem of transfer learning in text summarization.
We studied this problem from different perspectives through transfer of network layers from a pre-trained model to proposing a reinforcement learning framework which borrows insights from a self-critic policy gradient strategy and offers a systematic mechanism that creates a trade-off between the amount of reliance on the source or target dataset during training.
Through an extensive set of experiments, we demonstrated the generalization power of the proposed model on unseen test datasets and how it can reach state-of-the-art results on such datasets.
To the best of our knowledge, this is the first work that studies transfer learning in text summarization and offers a solution that beats state-of-the-art models and generalizes well to unseen datasets.
\vspace{-3mm}
\section*{Acknowledgments}
This work was supported in part by the US National Science Foundation grants IIS-1619028, IIS-1707498, and IIS-1838730.

\bibliographystyle{abbrv}
\vspace{-3mm}
\bibliography{reference}

\newpage

\section{Supplemental Material}
In this section, we provide the full details of our experimental analysis. Each table in this document represents a specific table in the main paper.

\begin{table*}[]
\centering
\label{table:trlresults}
\caption{(Table 3 in the main paper) Results on Newsroom, CNN/DM, DUC'03, and DUC'04 test data. $D_S$ shows the dataset that is used during pre-training and $D_G$ is our target dataset. \textbf{N} stands for Newsroom and \textbf{C} stands for CNN/DM dataset. The method column shows whether we use CE loss, transferring layers (TL), or TransferRL (TRL) loss during training. For each experiment, we run two different setups, with coverage mechanism and without it. This is represented as  We use coverage mechanism for all experiments. The result from the proposed method is shown with $\star$.}
\scriptsize
\begin{tabular}{|c|c|c|c|c|c|c|c|c|c|c|c|c|c|c|c|}
\hline
\multicolumn{1}{|l|}{\textbf{$D_M$}} & \multicolumn{1}{l|}{\textbf{$D_G$}} & \textbf{Method} & \textbf{Cov} & \multicolumn{3}{c|}{\textbf{Newsroom}}          & \multicolumn{3}{c|}{\textbf{CNN/DM}}                                                 & \multicolumn{3}{c|}{\textbf{DUC'03}}                                                & \multicolumn{3}{c|}{\textbf{DUC'04}}                                                \\ \hline
\multicolumn{4}{|l|}{\textbf{}}                                                                             & \textbf{$R_1$}             & \textbf{$R_2$}             & \textbf{$R_L$}             & \textbf{$R_1$}             & \textbf{$R_2$}             & \textbf{$R_L$}             & \textbf{$R_1$}             & \textbf{$R_2$}            & \textbf{$R_L$}             & \textbf{$R_1$}             & \textbf{$R_2$}            & \textbf{$R_L$}             \\ \hline
\multicolumn{2}{|c|}{\textbf{C+N}}                                         & CE Loss         & No           & \multicolumn{1}{l|}{31.22} & \multicolumn{1}{l|}{18.69} & \multicolumn{1}{l|}{27.94} & \multicolumn{1}{l|}{36.81} & \multicolumn{1}{l|}{15.44} & \multicolumn{1}{l|}{33.18} & \multicolumn{1}{l|}{27.58} & \multicolumn{1}{l|}{8.66} & \multicolumn{1}{l|}{24.61} & \multicolumn{1}{l|}{28.34} & \multicolumn{1}{l|}{8.86} & \multicolumn{1}{l|}{25.54} \\ \hline
\multicolumn{2}{|c|}{\textbf{C+N}}                                         & CE Loss         & Yes          & \multicolumn{1}{l|}{30.26} & \multicolumn{1}{l|}{17.68} & \multicolumn{1}{l|}{27.03} & \multicolumn{1}{l|}{38.23} & \multicolumn{1}{l|}{16.31} & \multicolumn{1}{l|}{34.66} & \multicolumn{1}{l|}{26.71} & \multicolumn{1}{l|}{7.81} & \multicolumn{1}{l|}{24.13} & \multicolumn{1}{l|}{27.96} & \multicolumn{1}{l|}{8.25} & \multicolumn{1}{l|}{25.25} \\ \hline\hline
\multicolumn{2}{|c|}{\textbf{N}}                                           & CE Loss         & No           & 36.08                      & 24.23                      & 32.79                      & 33.67                      & 12.79                      & 29.78                      & 28.22                      & 9.53                      & 24.99                      & 30.19                      & 10.44                     & 27.06                      \\ \hline
\multicolumn{2}{|c|}{\textbf{N}}                                           & CE Loss         & Yes          & 36.16                      & 24.33                      & 32.87                      & 33.58                      & 12.76                      & 29.72                      & 28.03                      & 9.15                      & 24.75                      & 29.85                      & 10.3                      & 26.7                       \\ \hline
\textbf{N}                           & \textbf{C}                          & TL              & No           & 31.43                      & 19.07                      & 28.13                      & 36.45                      & 14.87                      & 32.66                      & 27.69                      & 8.64                      & 24.54                      & 29.01                      & 9.19                      & 26.14                      \\ \hline
\textbf{N}                           & \textbf{C}                          & TL              & Yes          & 35.37                      & 23.45                      & 32.07                      & 34.51                      & 13.49                      & 30.61                      & 28.19                      & 9.34                      & 24.96                      & 29.83                      & 9.98                      & 26.66                      \\ \hline
\textbf{N}                           & \textbf{C}                          & TRL$^\star$     & No           & 36.04                      & 24.22                      & 32.78                      & 33.65                      & 12.76                      & 29.77                      & 28.32                      & 9.5                       & 25.12                      & 30.37                      & 10.58                     & 27.28                      \\ \hline
\textbf{N}                           & \textbf{C}                          & TRL$^\star$     & Yes          & \textbf{36.5}              & \textbf{24.77}             & \textbf{33.25}             & 35.24                      & 13.56                      & 31.33                      & \textbf{28.46}             & \textbf{9.65}             & \textbf{25.45}             & \textbf{30.45}             & \textbf{10.63}            & \textbf{27.42}             \\ \hline\hline
\multicolumn{2}{|c|}{\textbf{C}}                                           & CE Loss         & No           & 26.4                       & 13.25                      & 23.07                      & 39.11                      & 16.81                      & 35.64                      & 25.62                      & 6.68                      & 23.09                      & 26.25                      & 6.97                      & 23.76                      \\ \hline
\multicolumn{2}{|c|}{\textbf{C}}                                           & CE Loss         & Yes          & 26.59                      & 14.09                      & 23.44                      & 39.77                      & 17.15                      & 36.18                      & 25.66                      & 6.86                      & 23.16                      & 27.12                      & 7.07                      & 24.57                      \\ \hline
\textbf{C}                           & \textbf{N}                          & TL              & No           & 35.85                      & 24.06                      & 32.56                      & 34.39                      & 13.29                      & 30.49                      & 28.4                       & 9.63                      & 25.37                      & 30.13                      & 10.38                     & 27.06                      \\ \hline
\textbf{C}                           & \textbf{N}                          & TL              & Yes          & 35.39                      & 23.69                      & 32.13                      & 35.31                      & 13.88                      & 31.4                       & 28.34                      & 9.33                      & 25.09                      & 30.05                      & 10.1                      & 26.96                      \\ \hline
\textbf{C}                           & \textbf{N}                          & TRL$^\star$     & No           & 26.6                       & 13.41                      & 23.15                      & 39.18                      & 17.01                      & 35.72                      & 25.57                      & 6.7                       & 23.05                      & 26.31                      & 6.98                      & 23.85                      \\ \hline
\textbf{C}                           & \textbf{N}                          & TRL$^\star$     & Yes          & 27.34                      & 14.23                      & 24.01                      & \textbf{39.81}             & \textbf{17.23}             & \textbf{36.31}             & 25.61                      & 6.78                      & 23.18                      & 27.14                      & 7.11                      & 24.71                      \\ \hline
\end{tabular}
\end{table*}

\begin{table*}[]
\centering
\caption{(Table 4 in the main paper) Normalized and weighted normalized F-Scores for Table~\ref{table:trlresults}.}
\begin{tabular}{|c|c|c|c|c|c|c|c|c|c|}
\hline
\textbf{$D_S$} & \multicolumn{1}{l|}{\textbf{$D_G$}} & \textbf{Method} & \textbf{Cov} & \multicolumn{3}{c|}{\textbf{Avg. Score}}         & \multicolumn{3}{c|}{\textbf{\begin{tabular}[c]{@{}c@{}}Weighted\\ Avg. Score\end{tabular}}} \\ \hline
\multicolumn{4}{|l|}{}                                                                & \textbf{$R_1$} & \textbf{$R_2$} & \textbf{$R_L$} & \textbf{$R_1$}                & \textbf{$R_2$}               & \textbf{$R_L$}               \\ \hline
\multicolumn{2}{|c|}{\textbf{C+N}}                   & CE Loss         & No           & 30.99          & 12.91          & 27.82          & 32.47                         & 17.95                        & 29.11                        \\ \hline
\multicolumn{2}{|c|}{\textbf{C+N}}                   & CE Loss         & Yes          & 30.79          & 12.51          & 27.77          & 32.04                         & 17.36                        & 28.74                        \\ \hline\hline
\multicolumn{2}{|c|}{\textbf{N}}                     & CE Loss         & No           & 32.04          & 14.25          & 28.66          & 35.53                         & 21.66                        & 32.11                        \\ \hline
\multicolumn{2}{|c|}{\textbf{N}}                     & CE Loss         & Yes          & 31.91          & 14.14          & 28.51          & 35.58                         & 21.73                        & 32.16                        \\ \hline
\textbf{N}     & \textbf{C}                          & TL              & No           & 31.15          & 12.94          & 27.87          & 32.55                         & 18.12                        & 29.14                        \\ \hline
\textbf{N}     & \textbf{C}                          & TL              & Yes          & 31.98          & 14.07          & 28.58          & 35.17                         & 21.21                        & 31.74                        \\ \hline
\textbf{N}     & \textbf{C}                          & TRL$^\star$     & No           & 32.1           & 14.27          & 28.74          & 35.5                          & 21.64                        & 32.1                         \\ \hline
\textbf{N}     & \textbf{C}                          & TRL$^\star$     & Yes          & \textbf{32.66} & \textbf{14.65} & \textbf{29.36} & \textbf{36.21}                & \textbf{22.25}               & \textbf{32.81}               \\ \hline\hline
\multicolumn{2}{|c|}{\textbf{C}}                     & CE Loss         & No           & 29.35          & 10.93          & 26.39          & 29.25                         & 14.04                        & 25.89                        \\ \hline
\multicolumn{2}{|c|}{\textbf{C}}                     & CE Loss         & Yes          & 29.79          & 11.29          & 26.84          & 29.54                         & 14.77                        & 26.29                        \\ \hline
\textbf{C}     & \textbf{N}                          & TL              & No           & 32.26          & 14.34          & 28.87          & 35.52                         & 21.64                        & 32.09                        \\ \hline
\textbf{C}     & \textbf{N}                          & TL              & Yes          & 32.27          & 14.25          & 28.9           & 35.37                         & 21.48                        & 31.96                        \\ \hline
\textbf{C}     & \textbf{N}                          & TRL$^\star$     & No           & 29.42          & 11.03          & 26.44          & 29.42                         & 14.21                        & 25.97                        \\ \hline
\textbf{C}     & \textbf{N}                          & TRL$^\star$     & Yes          & 29.98          & 11.34          & 27.05          & 30.13                         & 14.9                         & 26.76                        \\ \hline
\end{tabular}
\end{table*}

\begin{table*}[]
\centering
\caption{(Table 5 in the main paper) Result of TransferRL after clipping $\zeta$ at 0.5 and $\zeta=1.0$ on Newsroom, CNN/DM, DUC'03, and DUC'04 test data along with the average and weighted average scores.}
\label{table:zetaclipping}
\scriptsize
\begin{tabular}{|c|c|c|c|c|c|c|c|c|c|c|c|c|c|c|c|c|}
\hline
\multicolumn{1}{|l|}{\textbf{$D_M$}} & \multicolumn{1}{l|}{\textbf{$D_G$}} & \multicolumn{1}{l|}{\textbf{$\zeta$}} & \textbf{Method} & \textbf{Cov} & \multicolumn{3}{c|}{\textbf{Newsroom}}          & \multicolumn{3}{c|}{\textbf{CNN/DM}}                                                 & \multicolumn{3}{c|}{\textbf{DUC'03}}                                                & \multicolumn{3}{c|}{\textbf{DUC'04}}                                                \\ \hline
\multicolumn{5}{|l|}{\textbf{}}                                                                             & \textbf{$R_1$}             & \textbf{$R_2$}             & \textbf{$R_L$}             & \textbf{$R_1$}             & \textbf{$R_2$}             & \textbf{$R_L$}             & \textbf{$R_1$}             & \textbf{$R_2$}            & \textbf{$R_L$}             & \textbf{$R_1$}             & \textbf{$R_2$}            & \textbf{$R_L$}             \\ \hline
\textbf{N}                           & \textbf{C}  & 1.0                        & TRL$^\star$     & No           & 36.04                      & 24.22                      & 32.78                      & 33.65                      & 12.76                      & 29.77                      & 28.32                      & 9.5                       & 25.12                      & 30.37                      & 10.58                     & 27.28                      \\ \hline
\textbf{N}                           & \textbf{C}  & 1.0                        & TRL$^\star$     & Yes          & \textbf{36.5}              & \textbf{24.77}             & \textbf{33.25}             & 35.24                      & 13.56                      & 31.33                      & \textbf{28.46}             & \textbf{9.65}             & \textbf{25.45}             & \textbf{30.45}             & \textbf{10.63}            & \textbf{27.42}             \\ \hline
\textbf{C}                           & \textbf{N}  & 1.0                         & TRL$^\star$     & No           & 26.6                       & 13.41                      & 23.15                      & 39.18                      & 17.01                      & 35.72                      & 25.57                      & 6.7                       & 23.05                      & 26.31                      & 6.98                      & 23.85                      \\ \hline
\textbf{C}                           & \textbf{N} & 1.0                          & TRL$^\star$     & Yes          & 27.34                      & 14.23                      & 24.01                      & \textbf{39.81}             & \textbf{17.23}             & \textbf{36.31}             & 25.61                      & 6.78                      & 23.18                      & 27.14                      & 7.11                      & 24.71                      \\ \hline\hline
\textbf{N}                           & \textbf{C}       & 0.5                   & TRL$^\star$     & No           & 36.2                       & 24.38                      & 32.92                      & 33.76                      & 12.86                      & 29.87                      & 28.38                      & 9.61                      & 25.12                      & 30.34                      & 10.56                     & 27.17                      \\ \hline
\textbf{N}                           & \textbf{C}       & 0.5                   & TRL$^\star$     & Yes          & 36.06                      & 24.23                      & 32.78                      & 33.7                       & 12.83                      & 29.81                      & 28.3                       & 9.54                      & 25.04                      & 29.88                      & 10.23                     & 26.8                       \\ \hline
\textbf{C}                           & \textbf{N}            & 0.5              & TRL$^\star$     & No           & 26.4                       & 13.25                      & 23.07                      & 39.1                       & 16.81                      & 35.63                      & 25.58                      & 6.69                      & 23.08                      & 26.22                      & 6.91                      & 23.75                      \\ \hline
\textbf{C}                           & \textbf{N}             & 0.5             & TRL$^\star$     & Yes          & 27.52                      & 14.46                      & 24.19                      & 38.82                      & 16.64                      & 35.22                      & 24.8                       & 6.49                      & 22.35                      & 25.65                      & 6.72                      & 23.19                      \\ \hline
\end{tabular}
\end{table*}

\begin{table*}[]
\centering
\caption{(Table 5 in the main paper) Normalized and weighted normalized F-Scores for Table~\ref{table:zetaclipping}.}
\begin{tabular}{|c|c|c|c|c|c|c|c|c|c|c|}
\hline
\textbf{$D_S$} & \multicolumn{1}{l|}{\textbf{$D_G$}} & \multicolumn{1}{l|}{\textbf{$\zeta$}} & \textbf{Method} & \textbf{Cov} & \multicolumn{3}{c|}{\textbf{Avg. Score}}         & \multicolumn{3}{c|}{\textbf{\begin{tabular}[c]{@{}c@{}}Weighted\\ Avg. Score\end{tabular}}} \\ \hline
\multicolumn{5}{|l|}{}                                                                & \textbf{$R_1$} & \textbf{$R_2$} & \textbf{$R_L$} & \textbf{$R_1$}                & \textbf{$R_2$}               & \textbf{$R_L$}               \\ \hline
\textbf{N}     & \textbf{C}     & 1.0                     & TRL$^\star$     & No           & 32.1           & 14.27          & 28.74          & 35.5                          & 21.64                        & 32.1                         \\ \hline
\textbf{N}     & \textbf{C}     & 1.0                     & TRL$^\star$     & Yes          & \textbf{32.66} & \textbf{14.65} & \textbf{29.36} & \textbf{36.21}                & \textbf{22.25}               & \textbf{32.81}               \\ \hline
\textbf{C}     & \textbf{N}     & 1.0                     & TRL$^\star$     & No           & 29.42          & 11.03          & 26.44          & 29.42                         & 14.21                        & 25.97                        \\ \hline
\textbf{C}     & \textbf{N}     & 1.0                     & TRL$^\star$     & Yes          & 29.98          & 11.34          & 27.05          & 30.13                         & 14.9                         & 26.76                        \\ \hline\hline
\textbf{N}     & \textbf{C}     & 0.5                     & TRL$^\star$     & No           & 32.17          & 14.36          & 28.77          & 35.65                         & 21.79                        & 32.23                        \\ \hline
\textbf{N}     & \textbf{C}     & 0.5                     & TRL$^\star$     & Yes          & 31.99          & 14.21          & 28.61          & 35.53                         & 21.66                        & 32.11                        \\ \hline
\textbf{C}     & \textbf{N}     & 0.5                     & TRL$^\star$     & No           & 29.33          & 10.92          & 26.38          & 29.24                         & 14.04                        & 25.88                        \\ \hline
\textbf{C}     & \textbf{N}     & 0.5                     & TRL$^\star$     & Yes          & 29.2           & 11.08          & 26.24          & 30.05                         & 14.94                        & 26.66                        \\ \hline
\end{tabular}
\end{table*}

\begin{table*}[!t]
\centering
\caption{(Table 6 in the main paper) Result of transfer learning methods using Newsroom for pre-training and DUC'03 for fine-tuning. The underlined result shows that the improvement from TL is not statistically significant compared to the proposed model.}
\label{table:duc03}
\begin{tabular}{|c|c|c|c|c|c|c|}
\hline
\textbf{$D_S$} & \textbf{$D_G$}    & \textbf{Method} & \textbf{Cov} & \textbf{$R_1$} & \textbf{$R_2$} & \textbf{$R_L$} \\ \hline
\textbf{N} & \textbf{DUC'03}        & TL    & No        & 28.76 &	9.73 &	25.54  \\ \hline
\textbf{N} & \textbf{DUC'03}        & TL    & Yes        & \underline{28.9}	& \underline{9.67}	& \underline{25.56}   \\ \hline
\textbf{N} & \textbf{DUC'03} & TRL$^\star$  & No         & 28.29 &	9.18	& 24.91          \\ \hline
\textbf{N} & \textbf{DUC'03} & TRL$^\star$  & Yes         & \underline{28.76} &	\underline{9.5} &	\underline{25.39}          \\ \hline\hline
\textbf{C} & \textbf{DUC'03}        & TL    & No        & 27.51	& 8.17 &	25.57          \\ \hline
\textbf{C} & \textbf{DUC'03}        & TL    & Yes        & 28.08 &	7.83 &	25.32          \\ \hline
\textbf{C} & \textbf{DUC'03} & TRL$^\star$  & No         & 24.94 &	6.21 &	22.44          \\ \hline
\textbf{C} & \textbf{DUC'03} & TRL$^\star$  & Yes         & 24.68 &	6.32 &	22.15          \\ \hline
\end{tabular}
\end{table*}

\begin{table*}[!t]
\centering
\caption{(Table 7 in the main paper) Result of transfer learning methods using Newsroom for pre-training and DUC'04 for fine-tuning.}
\label{table:duc03}
\begin{tabular}{|c|c|c|c|c|c|c|}
\hline
\textbf{$D_S$} & \textbf{$D_G$}    & \textbf{Method} & \textbf{Cov} & \textbf{$R_1$} & \textbf{$R_2$} & \textbf{$R_L$} \\ \hline
\textbf{N} & \textbf{DUC'04}        & TL    & No        & 30.22 &	10.55	& 27.15 \\ \hline
\textbf{N} & \textbf{DUC'04}        & TL    & Yes        & 27.68 &	9.09	& 25.29          \\ \hline
\textbf{N} & \textbf{DUC'04} & TRL$^\star$  & No         & \textbf{30.35} &	\textbf{10.43}	& \textbf{27.32}          \\ \hline
\textbf{N} & \textbf{DUC'04} & TRL$^\star$  & Yes         & 29.54 &	9.99 &	26.56          \\ \hline\hline
\textbf{C} & \textbf{DUC'04}        & TL    & No        & 28.45 &	8.41 &	26.75          \\ \hline
\textbf{C} & \textbf{DUC'04}        & TL    & Yes        & 28.45 &	8.05 &	26.54          \\ \hline
\textbf{C} & \textbf{DUC'04} & TRL$^\star$  & No         & 25.98 &	6.71 &	23.51          \\ \hline
\textbf{C} & \textbf{DUC'04} & TRL$^\star$  & Yes         & 26.05 &	6.63 &	23.54          \\ \hline
\end{tabular}
\end{table*}




\end{document}